\documentclass[runningheads]{llncs}

\usepackage{graphicx, amsmath, amssymb, caption, subcaption, multirow, overpic, textpos, multibib}
\usepackage[table]{xcolor}

\usepackage{graphicx}

\usepackage{tikz}
\usepackage{comment}
\usepackage{amsmath,amssymb} %
\usepackage{color}
\usepackage{enumitem}

\usepackage[accsupp]{axessibility}  %

\usepackage[british,english,american]{babel}
\definecolor{citecolor}{HTML}{0071BC}
\definecolor{linkcolor}{HTML}{ED1C24}
\newcommand{\app}{\raise.17ex\hbox{$\scriptstyle\sim$}}

\usepackage[pagebackref=false, breaklinks=true, letterpaper=true, colorlinks, citecolor=citecolor, linkcolor=linkcolor, bookmarks=false]{hyperref}

\usepackage{xspace}
\makeatletter
\DeclareRobustCommand\onedot{\futurelet\@let@token\@onedot}
\def\@onedot{\ifx\@let@token.\else.\null\fi\xspace}

\def\eg{\emph{e.g}\onedot} 
\def\ie{\emph{i.e}\onedot} 
 
\def\etc{\emph{etc}\onedot} \def\vs{\emph{vs}\onedot}

\makeatother

\newcolumntype{x}[1]{>{\centering\arraybackslash}p{#1pt}}
\newcolumntype{y}[1]{>{\raggedright\arraybackslash}p{#1pt}}
\newcolumntype{z}[1]{>{\raggedleft\arraybackslash}p{#1pt}}
\newlength\savewidth\newcommand\shline{\noalign{\global\savewidth\arrayrulewidth
  \global\arrayrulewidth 1pt}\hline\noalign{\global\arrayrulewidth\savewidth}}

\newcommand{\tablestyle}[2]{\ttfamily\setlength{\tabcolsep}{#1}\renewcommand{\arraystretch}{#2}\centering\footnotesize}

\definecolor{gain}{HTML}{34a853}  %
\newcommand{\gain}[1]{\textcolor{gain}{#1}}
\definecolor{lost}{HTML}{ea4335}  %
\newcommand{\lost}[1]{\textcolor{lost}{#1}}

\newcites{app}{Appendix References}

\newcommand{\res}[2]{{#1} {({\gain{#2}})}}

\makeatletter\renewcommand\paragraph{\@startsection{paragraph}{4}{\z@}
  {.5em \@plus1ex \@minus.2ex}{-.5em}{\normalfont\normalsize\bfseries}}\makeatother

\definecolor{baselinecolor}{gray}{.9}

\newcommand{\boxAP}{AP$^\text{box}$\xspace}
\newcommand{\maskAP}{AP$^\text{mask}$\xspace}
\newcommand{\maskAPrare}{AP$^\text{mask}_\text{rare}$\xspace}

\begin{document}
\pagestyle{headings}
\mainmatter

\title{Exploring Plain Vision Transformer Backbones \\ for Object Detection \vspace{-.5em}}

\titlerunning{~}
\author{
Yanghao Li \quad
Hanzi Mao \quad
Ross Girshick$^\dagger$ \quad
Kaiming He$^\dagger$
\\{\scriptsize $^\dagger$equal contribution}
}
\authorrunning{~}
\institute{Facebook AI Research}
\maketitle

\begin{abstract}
\vspace{-2em}
We explore the \textit{plain}, \textit{non-hierarchical} Vision Transformer (ViT) as a backbone network for object detection. This design enables the original ViT architecture to be fine-tuned for object detection without needing to redesign a hierarchical backbone for pre-training. With minimal adaptations for fine-tuning, our plain-backbone detector can achieve competitive results. Surprisingly, we observe: (i) it is sufficient to build a simple feature pyramid from a single-scale feature map (without the common FPN design) and (ii) it is sufficient to use window attention (without shifting) aided with very few cross-window propagation blocks. With plain ViT backbones pre-trained as Masked Autoencoders (MAE), our detector, named \mbox{ViTDet}, can compete with the previous leading methods that were all based on hierarchical backbones, reaching up to 61.3 AP$^\text{box}$ on the COCO dataset using only ImageNet-1K pre-training. We hope our study will draw attention to research on plain-backbone detectors. Code for ViTDet is available.\footnote{\url{https://github.com/facebookresearch/detectron2/tree/main/projects/ViTDet}}
\vspace{-2em}
\end{abstract}

\begin{figure}[t]
\vspace{-.5em}
\centering
\includegraphics[width=1.0\linewidth]{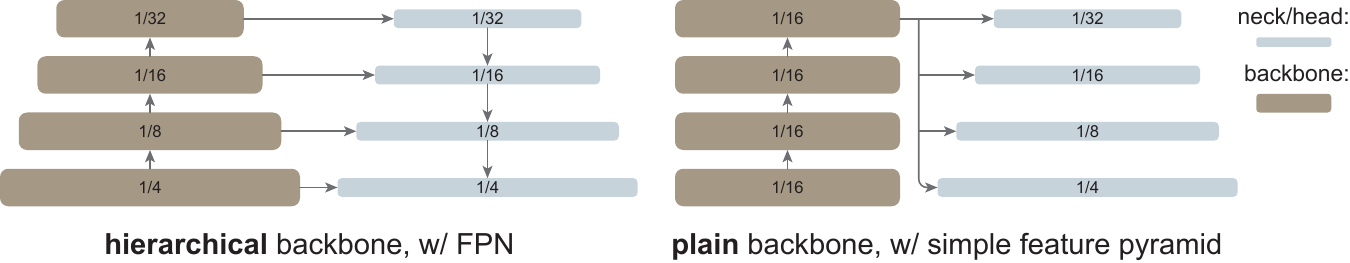}
\vspace{-1.2em}
\caption{A typical hierarchical-backbone detector (left) \vs our plain-backbone detector (right).
Traditional  hierarchical backbones can be naturally adapted for multi-scale detection, \eg, using FPN. Instead, we explore building a simple pyramid from only the last, large-stride (16) feature map of a plain backbone.
}
\label{fig:arch}
\vspace{-.75em}
\end{figure}

\section{Introduction}

Modern object detectors in general consist of a \textit{backbone} feature extractor that is \textit{agnostic} to the detection task and a set of necks and heads that incorporate detection-specific prior knowledge. Common components in the necks/heads may include Region-of-Interest (RoI) operations \cite{He2014,Girshick2015,He2017}, Region Proposal Networks (RPN) or anchors \cite{Ren2015}, Feature Pyramid Networks (FPN) \cite{Lin2017}, \etc. If the design of the task-specific necks/heads is decoupled from the design of the backbone, they may evolve in parallel. Empirically, object detection research has benefited from the largely independent exploration of general-purpose backbones \cite{Krizhevsky2012,Simonyan2015,Szegedy2015,He2016} and detection-specific modules. For a long while, these backbones have been \textit{multi-scale}, \textit{hierarchical} architectures due to the \textit{de facto} design of convolutional networks (ConvNet) \cite{LeCun1989}, which has heavily influenced the neck/head design for detecting objects at multiple scales (\eg, FPN).

Over the past year, Vision Transformers (ViT) \cite{Dosovitskiy2021} have been established as a powerful backbone for visual recognition. Unlike typical ConvNets, the original ViT is a \textit{plain, non-hierarchical} architecture that maintains a single-scale feature map throughout. 
Its ``minimalist" pursuit is met by challenges when applied to object detection---\eg, How can we address multi-scale objects in a downstream task with a plain backbone from upstream pre-training? Is a plain ViT too inefficient to use with high-resolution detection images? One solution, which abandons this pursuit, is to re-introduce hierarchical designs into the backbone. This solution, \eg, Swin Transformers \cite{Liu2021} and related works \cite{Wang2021,Fan2021,Li2021a,Heo2021}, can inherit the ConvNet-based detector design and has shown successful results.

In this work, we pursue a different direction: we explore object detectors that use only \textit{plain, non-hierarchical} backbones.\footnotemark~If this direction is successful, it will enable the use of original ViT backbones for object detection; this will \textit{decouple} the pre-training design from the fine-tuning demands, maintaining the independence of upstream \vs downstream tasks, as has been the case for ConvNet-based research. This direction also in part follows the ViT philosophy of ``fewer inductive biases" \cite{Dosovitskiy2021} in the pursuit of universal features. 
As the non-local self-attention computation \cite{Vaswani2017} can learn translation-equivariant features \cite{Dosovitskiy2021}, they may also learn scale-equivariant features from certain forms of supervised or self-supervised pre-training.

\footnotetext{In this paper, ``backbone'' refers to architectural components that can be inherited from pre-training and ``plain'' refers to the \mbox{non-hierarchical}, single-scale property.}

In our study, we do \textit{not} aim to develop new components; instead, we make \textit{minimal} adaptations that are sufficient to overcome the aforementioned challenges. In particular, our detector builds a simple feature pyramid from only the \textit{last} feature map of a plain ViT backbone (Figure~\ref{fig:arch}). This abandons the FPN design \cite{Lin2017} and waives the requirement of a hierarchical backbone. To efficiently extract features from high-resolution images, our detector uses simple non-overlapping window attention (without ``shifting", unlike \cite{Liu2021}). A small number of cross-window blocks (\eg, 4), which could be global attention \cite{Vaswani2017} or convolutions, are used to propagate information. These adaptations are made only during fine-tuning and do not alter pre-training.

Our simple design turns out to achieve surprising results. We find that the FPN design is not necessary in the case of a plain ViT backbone and its benefit can be effectively gained by a simple pyramid built from a large-stride (16), single-scale  map. We also find that window attention is sufficient as long as information is well propagated across windows in a small number of layers.

More surprisingly, under some circumstances, our plain-backbone detector, named {ViTDet}, can compete with the leading hierarchical-backbone detectors (\eg, Swin \cite{Liu2021}, MViT \cite{Fan2021,Li2021a}).
With Masked Autoencoder (MAE) \cite{He2021} pre-training, our plain-backbone detector can outperform the hierarchical counterparts that are pre-trained on ImageNet-1K/21K \cite{Deng2009} with supervision (Figure~\ref{fig:tradeoff}).
The gains are more prominent for larger model sizes. 
The competitiveness of our detector is observed under different object detector frameworks, including Mask R-CNN \cite{He2017}, Cascade Mask R-CNN \cite{Cai2019}, and their enhancements.
We report 61.3 \boxAP on the COCO dataset \cite{Lin2014} with a plain ViT-Huge backbone, using only ImageNet-1K pre-training with no labels. We also demonstrate competitive results on the long-tailed LVIS detection dataset \cite{Gupta2019}.
While these strong results may be in part due to the effectiveness of MAE pre-training, our study demonstrates that plain-backbone detectors can be promising, challenging the entrenched position of hierarchical backbones for object detection.

Beyond these results, our methodology maintains the philosophy of decoupling the detector-specific designs from the task-agnostic backbone. This philosophy is in contrast to the trend of redesigning Transformer backbones to support multi-scale hierarchies \cite{Liu2021,Wang2021,Fan2021,Heo2021}. In our case, the detection-specific prior knowledge is introduced only during fine-tuning, without needing to tailor the backbone design a priori in pre-training. This makes our detector compatible with ViT developments along various directions that are not necessarily limited by the hierarchical constraint, \eg, block designs \cite{Tolstikhin2021,Touvron2021c}, self-supervised learning \cite{Bao2021,He2021}, and scaling \cite{Zhai2021}. We hope our study will inspire future research on plain-backbone object detection.\footnotemark

\footnotetext{This work is an extension of a preliminary version \cite{Li2021b} that was unpublished and not submitted for peer review.}

\section{Related Work} \label{sec:related}

\paragraph{Object detector backbones.} Pioneered by the work of R-CNN \cite{Girshick2014}, object detection and many other vision tasks adopt a pre-training + fine-tuning paradigm: a general-purpose, task-agnostic backbone is pre-trained with supervised or self-supervised training, whose structure is later modified and adapted to the downstream tasks. The dominant backbones in computer vision have been ConvNets \cite{LeCun1989} of various forms, \eg, \cite{Krizhevsky2012,Simonyan2015,Szegedy2015,He2016}.

Earlier neural network detectors, \eg, \cite{He2014,Girshick2015,Ren2015,Redmon2016}, were based on a single-scale feature map when originally presented. While they use ConvNet backbones that are by default hierarchical, in principle, they are applicable on any plain backbone. SSD \cite{Liu2016} is among the first works that leverage the hierarchical nature of the ConvNet backbones (\eg, the last two stages of a VGG net \cite{Simonyan2015}). FPN \cite{Lin2017} pushes this direction further by using all stages of a hierarchical backbone, approached by lateral and top-down connections. The FPN design is widely used in object detection methods. More recently, works including Trident Networks~\cite{li2019scale} and YOLOF~\cite{chen2021you} have revisited single-scale feature maps, but unlike our work they focus on a single-scale taken from a \emph{hierarchical} backbone.

ViT \cite{Dosovitskiy2021} is a powerful alternative to standard ConvNets for image classification. The original ViT is a plain, non-hierarchical architecture. Various hierarchical Transformers have been presented, \eg, Swin \cite{Liu2021}, MViT \cite{Fan2021,Li2021a}, PVT \cite{Wang2021}, and PiT \cite{Heo2021}. These methods inherit some designs from ConvNets, including the hierarchical structure and the translation-equivariant priors (\eg, convolutions, pooling, sliding windows). As a result, it is relatively straightforward to replace a ConvNet with these backbones for object detection.

\paragraph{Plain-backbone detectors.} The success of ViT has inspired people to push the frontier of plain backbones for object detection. Most recently, UViT \cite{Chen2021b} is presented as a single-scale Transformer for object detection.
UViT studies the network width, depth, and input resolution of plain ViT backbones under object detection metrics. A progressive window attention strategy is proposed to address the high-resolution inputs.
Unlike UViT that modifies the architecture \textit{during pre-training}, our study focuses on the original ViT architecture \textit{without} a priori specification for detection. By maintaining the task-agnostic nature of the backbone, our approach supports a wide range of available ViT backbones as well as their improvements in the future. Our method \textit{decouples} the backbone design from the detection task, which is a key motivation of pursuing plain backbones.

UViT uses single-scale feature maps for the detector heads, while our method builds a simple pyramid on the single-scale backbone. In the context of our study, it is an unnecessary constraint for the entire detector to be single-scale. Note the full UViT detector has several forms of multi-scale priors too (\eg, RPN \cite{Ren2015} and RoIAlign \cite{He2017}) as it is based on Cascade Mask R-CNN \cite{Cai2019}.
In our study, we focus on leveraging pre-trained plain backbones and we do not constrain the detector neck/head design.

\paragraph{Object detection methodologies.} Object detection is a flourishing research area that has embraced methodologies of distinct properties---\eg, two-stage \cite{Girshick2014,He2014,Girshick2015,Ren2015} \vs one-stage \cite{Redmon2016,Liu2016,Lin2017a}, anchor-based \cite{Ren2015} \vs anchor-free \cite{Law2018,Duan2019,Tian2019a}, and region-based \cite{Girshick2014,He2014,Girshick2015,Ren2015} \vs query-based (DETR) \cite{Carion2020}. Research on different methodologies has been continuously advancing understandings of the object detection problem. Our study suggests that the topic of ``plain \vs hierarchical" backbones is worth exploring and may bring in new insights.

\section{Method}\label{sec:method}

Our goal is to remove the hierarchical constraint on the backbone and to enable explorations of plain-backbone object detection. To this end, we aim for \textit{minimal} modifications to adapt a plain backbone to the object detection task \textit{only during fine-tuning time}.
After these adaptations, in principle one can apply any detector heads, for which we opt to use Mask R-CNN \cite{He2017} and its extensions. We do \textit{not} aim to develop new components; instead, we focus on what new insights can be drawn in our exploration.

\begin{figure}[t]
    \centering
    \includegraphics[width=1.0\linewidth]{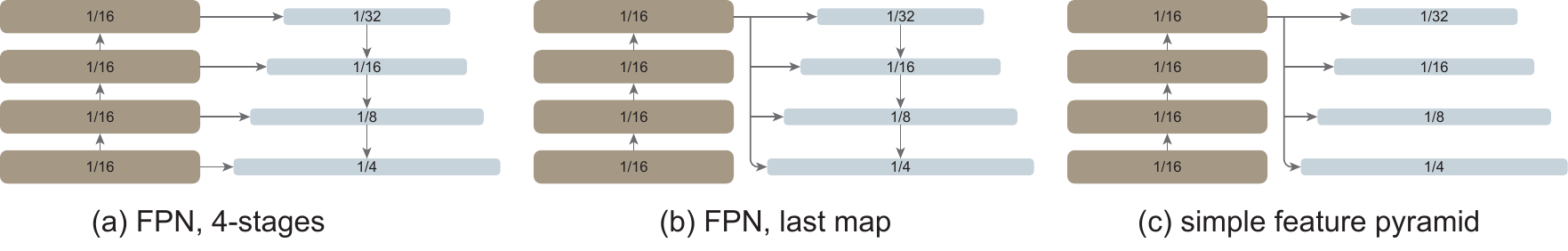}
    \vspace{-1.5em}
    \caption{Building a feature pyramid on a plain backbone. \textbf{(a)} FPN-like: to mimic a hierarchical backbone, the plain backbone is artificially divided into multiple stages. \textbf{(b)} FPN-like, but using only the last feature map without stage division. \textbf{(c)} Our simple feature pyramid without FPN. In all three cases, strided convolutions/deconvolutions are used whenever the scale changes. 
    }
    \label{fig:fpn}
\end{figure}

\paragraph{Simple feature pyramid.}

FPN \cite{Lin2017} is a common solution of building an \mbox{in-network} pyramid for object detection. If the backbone is hierarchical, the motivation of FPN is to combine the higher-resolution features from earlier stages and the stronger features from later stages. This is realized in FPN by top-down and lateral connections \cite{Lin2017} (Figure~\ref{fig:arch} left).

If the backbone is non-hierarchical, the foundation of the FPN motivation is lost, as all the feature maps in the backbone are of the same resolution. In our scenario, we simply use only the \textit{last} feature map from the backbone, which should have the strongest features. On this map, we apply a set of convolutions or deconvolutions \textit{in parallel} to produce multi-scale feature maps. Specifically, with the default ViT feature map of a scale of $\frac{1}{16}$ (stride = 16 \cite{Dosovitskiy2021}), we produce feature maps of scales $\{\frac{1}{32}, \frac{1}{16}, \frac{1}{8}, \frac{1}{4}\}$ using convolutions of strides $\{2, 1, \frac{1}{2}, \frac{1}{4}\}$, where a fractional stride indicates a deconvolution. We refer to this as a ``\textit{simple feature pyramid}" (Figure~\ref{fig:arch} right).

The strategy of building multi-scale feature maps from a single map is related to that of SSD \cite{Liu2016}. However, our scenario involves \textit{upsampling} from a deep, low-resolution feature map, unlike \cite{Liu2016}, which taps into shallower feature maps. In hierarchical backbones, upsampling is often aided by lateral connection \cite{Lin2017}; in plain ViT backbones, we empirically find this is not necessary (Sec.~\ref{sec:exp}) and simple deconvolutions are sufficient.
We hypothesize that this is because ViT can rely on positional embedding \cite{Vaswani2017} for encoding locations and also because the high-dimensional ViT patch embeddings do not necessarily discard information.\footnotemark

\footnotetext{With a patch size of 16$\times$16 and 3 colors, a hidden dimension $\ge$768 (ViT-B and larger) can preserve all information of a patch if necessary.}

We will compare with two FPN variants that are also built on a plain backbone (Figure~\ref{fig:fpn}). In the first variant, the backbone is artificially divided into multiple stages to mimic the stages of a hierarchical backbone, with lateral and top-down connections applied (Figure~\ref{fig:fpn} (a))~\cite{Elnouby2021}. The second variant is like the first one, but uses only the last map instead of the divided stages (Figure~\ref{fig:fpn} (b)). We show that these FPN variants are not necessary (Sec.~\ref{sec:exp}).\footnotemark

\footnotetext{From a broader perspective, the spirit of FPN \cite{Lin2017} is ``to build a feature pyramid inside a network". Our simple feature pyramid follows this spirit. In the context of this paper, the term of ``FPN" refers to the specific architecture design in \cite{Lin2017}.
}

\paragraph{Backbone adaptation.}

Object detectors benefit from high-resolution input images, but computing global self-attention throughout the backbone is prohibitive in memory and is slow. In this study, we focus on the scenario where the pre-trained backbone performs global self-attention, which is then \textit{adapted} to higher-resolution inputs during fine-tuning. This is in contrast to the recent methods that modify the attention computation directly with backbone pre-training (\eg, \cite{Liu2021,Fan2021}). Our scenario enables us to use the original ViT backbone for detection, without redesigning pre-training architectures.

We explore using \textit{window attention} \cite{Vaswani2017} with a few cross-window blocks. During fine-tuning, given a high-resolution feature map, we divide it into regular non-overlapping windows.\footnotemark~Self-attention is computed within each window. This is referred to as ``\textit{restricted}" self-attention in the original Transformer \cite{Vaswani2017}. 

\footnotetext{We set the window size as the pre-training feature map size by default (14$\times$14 \cite{Dosovitskiy2021}).}

Unlike Swin, we do \textit{not} ``shift" \cite{Liu2021} the windows across layers. To allow information propagation, we use a very few (by default, 4) blocks that can go across windows.
We \textit{evenly} split a pre-trained backbone into 4 subsets of blocks (\eg, 6 in each subset for the 24-block ViT-L). We apply a propagation strategy in the last block of each subset.
We study these two strategies:

\vspace{.5em}
(i)~\emph{Global propagation}. We perform global self-attention in the last block of each subset. As the number of global blocks is small, the memory and computation cost is feasible. This is similar to the hybrid window attention in~\cite{Li2021a} that was used jointly with FPN.

(ii)~\emph{Convolutional propagation}. As an alternative, we add an extra convolutional block after each subset.
A convolutional block is a residual block \cite{He2016} that consists of one or more convolutions and an identity shortcut. The last layer in this block is initialized as zero, such that the initial status of the block is an identity \cite{Goyal2017}. Initializing a block as identity allows us to insert it into any place in a pre-trained backbone without breaking the initial status of the backbone.

\vspace{.5em}
Our backbone adaptation is simple and makes detection fine-tuning compatible with global self-attention pre-training. As stated, it is not necessary to redesign the pre-training architectures.

\paragraph{Discussion.} Object detectors contain components that can be task agnostic, such as the backbone, and other components that are task-specific, such as RoI heads. This model decomposition enables the task-agnostic components to be pre-trained using non-detection data (\eg, ImageNet), which may provide an advantage since detection training data is relatively scarce.

Under this perspective, it becomes reasonable to pursue a backbone that involves fewer inductive biases, since the backbone may be trained effectively using large-scale data and/or self-supervision. In contrast, the detection task-specific components have relatively little data available and may still benefit from additional inductive biases. While pursuing detection heads with fewer inductive biases is an active area of work, leading methods like DETR~\cite{Carion2020} are challenging to train and still benefit from detection-specific prior knowledge \cite{Zhu2020}.

Driven by these observations, our work follows the spirit of the original plain ViT paper with respect to the detector's backbone. While the ViT paper's discussion \cite{Dosovitskiy2021} focused on reducing inductive biases on translation equivariance, in our case, it is about having fewer or even no inductive bias on scale equivariance in the backbone. We hypothesize that the way for a plain backbone to achieve scale equivariance is to learn the prior knowledge from data, analogous to how it learns translation equivariance and locality without convolutions \cite{Dosovitskiy2021}.

Our goal is to demonstrate the feasibility of this approach. Thus we choose to implement our method with standard detection specific components (\ie, Mask R-CNN and its extensions). Exploring even fewer inductive biases in the detection heads is an open and interesting direction for future work. We hope it can benefit from and build on our work here.

\paragraph{Implementation.} We use the vanilla ViT-B, ViT-L, ViT-H \cite{Dosovitskiy2021} as the pre-training backbones. We set the patch size as 16 and thus the feature map scale is 1/16, \ie, stride = 16.\footnotemark~Our detector heads follow Mask R-CNN~\cite{He2017} or Cascade Mask R-CNN~\cite{Cai2019}, with architectural details described in the appendix.
The input image is 1024$\times$1024, augmented with large-scale jittering \cite{Ghiasi2021} during training.
Due to this heavy regularization, we fine-tune for up to 100 epochs in COCO.
We use the AdamW optimizer~\cite{Loshchilov2019} and search for optimal hyper-parameters using a baseline version. More details are in the appendix.

\footnotetext{Changing the stride affects the scale distribution and presents a different accuracy shift for objects of different scales. This topic is beyond the scope of this study. For simplicity, we use the same patch size of 16 for all of ViT-B, L, H (see the appendix).
}

\section{Experiments} \label{sec:exp}

\subsection{Ablation Study and Analysis} \label{subsec:ablation}

We perform ablation experiments on the COCO dataset \cite{Lin2014}. We train on the \textsf{train2017} split and evaluate on the \textsf{val2017} split. We report results on bounding-box object detection (AP$^\text{box}$) and instance segmentation (AP$^\text{mask}$).

By default, we use the simple feature pyramid and global propagation described in Sec.~\ref{sec:method}. We use 4 propagation blocks, evenly placed in the backbone. We initialize the backbone with MAE \cite{He2021} pre-trained on IN-1K without labels. We ablate these defaults and discuss our main observations as follows.

\paragraph{A simple feature pyramid is sufficient.} In Table~\ref{tab:feature_pyramids} we compare the feature pyramid building strategies illustrated in Figure~\ref{fig:fpn}.

We study a baseline with \textit{no feature pyramid}: both the RPN and RoI heads are applied on the backbone's final, single-scale ($\frac{1}{16}$) feature map. This case is similar to the original Faster R-CNN \cite{Ren2015} before FPN was proposed. \textit{All} feature pyramid variants (Table~\ref{tab:feature_pyramids} a-c) are substantially better than this baseline, increasing AP by up to 3.4 points. We note that using a single-scale feature map does \textit{not} mean the detector is single-scale: the RPN head has multi-scale anchors and the RoI heads operate on regions of multiple scales. Even so, feature pyramids are beneficial. This observation is consistent with the observation in the FPN paper \cite{Lin2017} on hierarchical backbones.

However, \textit{the FPN design is not needed and our simple feature pyramid is sufficient} for a plain ViT backbone to enjoy the benefit of a pyramid. To ablate this design, we mimic the FPN architecture (\ie, the top-down and lateral connections) as in Figure~\mbox{\ref{fig:fpn} (a, b)}. 
Table~\ref{tab:feature_pyramids} (a, b) shows that
while both FPN variants achieve strong gains over the baseline with no pyramid (as has been widely observed with the original FPN on hierarchical backbones), they are no better than our simple feature pyramid.
The original FPN \cite{Lin2017} was motivated by combining lower-resolution, stronger feature maps with higher-resolution, weaker feature maps. This foundation is lost when the backbone is plain and has no high-resolution maps, which can explain why our simple pyramid is sufficient.

\begin{table}[t]
    \tablestyle{8pt}{1.1}
    \begin{tabular}{l|ll|ll}
     & \multicolumn{2}{c|}{ViT-B} & \multicolumn{2}{c}{ViT-L} \vspace{-.5em} \\
    pyramid design & \multicolumn{1}{c}{\scriptsize \boxAP} & \multicolumn{1}{c|}{\scriptsize \maskAP} & \multicolumn{1}{c}{\scriptsize \boxAP} & \multicolumn{1}{c}{\scriptsize \maskAP} \\
    \shline
     no feature pyramid & {47.8} & {42.5} & {51.2} & {45.4} \\
     \hline
     (a) FPN, 4-stage & \res{50.3}{+2.5} & \res{44.9}{+2.4} & \res{54.4}{+3.2} & \res{48.4}{+3.0} \\
     (b) FPN, last-map & \res{{50.9}}{+3.1} & \res{45.3}{+2.8} & \res{\textbf{54.6}}{+3.4} & \res{48.5}{+3.1} \\
     (c) simple feature pyramid & \res{\textbf{51.2}}{+3.4} & \res{\textbf{45.5}}{+3.0} & \res{\textbf{54.6}}{+3.4} & \res{\textbf{48.6}}{+3.2} \\
    \end{tabular}
    \vspace{1em}
    \caption{\textbf{Ablation on feature pyramid design} with plain ViT backbones, using {Mask R-CNN} evaluated on COCO. The backbone is \mbox{ViT-B} (left) and \mbox{ViT-L} (right).
    The entries (a-c) correspond to Figure~\ref{fig:fpn}~(a-c), compared to a baseline without any pyramid.
    Both FPN and our simple pyramid are substantially better than the baseline, while our simple pyramid is sufficient.
    \label{tab:feature_pyramids}
    }
\vspace{-1em}
\end{table}

Our ablation reveals that the \textit{set} of pyramidal feature maps, rather than the top-down/lateral connections, is the key to effective multi-scale detection. To see this, we study an even more aggressive case of the simple pyramid: we generate only the finest scale ($\frac{1}{4}$) feature map by deconvolution and then from this finest map we subsample other scales in parallel by strided \textit{average pooling}. There are no unshared, per-scale parameters in this design.
This aggressively-simple pyramid is nearly as good: it has 54.5 AP (ViT-L), 3.3 higher than the no pyramid baseline.
This shows the importance of pyramidal feature maps.
For any variant of these feature pyramids, the anchors (in RPN) and regions (in RoI heads) are mapped to the corresponding level in the pyramid based on their scales, as in \cite{Lin2017}. We hypothesize that this explicit scale-equivariant mapping, rather than the top-down/lateral connection, is the main reason why a feature pyramid can greatly benefit multi-scale object detection.

\paragraph{Window attention is sufficient when aided by a few propagation blocks.} Table~\ref{tab:backbone_ablations} ablates our backbone adaptation approach. In short, on top of a baseline that has purely window attention and none of the cross-window propagation blocks (Table~\ref{tab:backbone_ablations}, ``none"), various ways of propagation can show decent gains.\footnotemark

\footnotetext{Even our baseline with no propagation \emph{in the backbone} is reasonably good (52.9 AP). This can be explained by the fact that the layers beyond the backbone (the simple feature pyramid, RPN, and RoI heads) also induce cross-window communication.}

\newcommand{\bad}[2]{{#1} {({\lost{--#2}})}}
\begin{table}[t]
\centering
\subfloat[
Window attention with various cross-window propagation strategies.
\label{tab:backbone_ablation:prop}
]{
\centering
\begin{minipage}{0.46\linewidth}{\begin{center}
\tablestyle{4pt}{1.1}
\begin{tabular}{@{}y{60}|y{42}y{42}l@{}}
   prop. strategy & \multicolumn{1}{c}{AP$^\text{box}$} & \multicolumn{1}{c}{AP$^\text{mask}$} \\
    \shline
    none & {52.9} & 47.2 \\ 
    \hline
    4 global blocks & \res{54.6}{+1.7} & \res{48.6}{+1.4} \\
    4 conv blocks & \res{\textbf{54.8}}{+1.9} & \res{\textbf{48.8}}{+1.6} \\
    shifted win. & \res{54.0}{+1.1} & \res{47.9}{+0.7} \\
\end{tabular}
\end{center}}\end{minipage}
}
\hspace{1em}
\subfloat[
Convolutional propagation with different residual block types (4 blocks).
\label{tab:backbone_ablation:conv_type}
]{
\begin{minipage}{0.46\linewidth}{\begin{center}
\tablestyle{4pt}{1.1}
\begin{tabular}{@{}y{40}|y{42}y{42}@{}}
    prop. conv &  \multicolumn{1}{c}{AP$^\text{box}$} & \multicolumn{1}{c}{AP$^\text{mask}$} \\
    \shline
    none & 52.9 & 47.2\\
    \hline
    na\"ive & \res{54.3}{+1.4} & \res{48.3}{+1.1} \\
    basic & \res{\textbf{54.8}}{+1.9} &	\res{\textbf{48.8}}{+1.6}\\
    bottleneck & \res{54.6}{+1.7} & \res{48.6}{+1.4}\\ 
    \end{tabular}
\end{center}}\end{minipage}
}
\\
\subfloat[Locations of cross-window global propagation blocks.
\label{tab:backbone_ablation:block_place}
]{
\begin{minipage}{0.46\linewidth}{\begin{center}
\tablestyle{4pt}{1.1}
\begin{tabular}{@{}y{60}|y{42}y{42}@{}}
    prop. locations & \multicolumn{1}{c}{AP$^\text{box}$} & \multicolumn{1}{c}{AP$^\text{mask}$} \\
    \shline
    none & 52.9 & 47.2\\
    \hline
    first 4 blocks & {52.9} (+0.0) & {47.1} (--0.1)  \\
    last 4 blocks & \res{54.3}{+1.4} & \res{48.3}{+1.1} \\
    evenly 4 blocks & \res{\textbf{54.6}}{+1.7} & \res{\textbf{48.6}}{+1.4} \\
    \end{tabular}
\end{center}}\end{minipage}
}
\hspace{1em}
\subfloat[Number of global propagation blocks. $^\dagger$: Memory optimization required.
\label{tab:backbone_ablation:block_num}
]{
\begin{minipage}{0.46\linewidth}{\begin{center}
\tablestyle{4pt}{1.1}
\begin{tabular}{@{}y{40}|y{42}y{42}@{}}
    prop. blks &  \multicolumn{1}{c}{AP$^\text{box}$} & \multicolumn{1}{c}{AP$^\text{mask}$} \\
    \shline
    none & 52.9 & 47.2\\
    \hline
    2 & \res{54.4}{+1.5} & \res{48.5}{+1.3} \\
    4 & \res{54.6}{+1.7} & \res{48.6}{+1.4} \\
    24$^\dagger$ & \res{\textbf{55.1}}{+2.2} & \res{\textbf{48.9}}{+1.7} \\ 
    \end{tabular}
\end{center}}\end{minipage}
}
\vspace{-.5em}
\caption{\textbf{Ablation on backbone adaptation strategies} using a plain ViT backbone and {Mask R-CNN} evaluated on COCO. All blocks perform window attention, unless modified by the propagation strategy. In sum, compared to the baseline that uses only window attention (52.9 AP$^\text{box}$) most configurations work effectively as long as information can be well propagated across windows.
Here the backbone is ViT-L; the observations on ViT-B are similar (see the appendix).
}
\label{tab:backbone_ablations}
\vspace{-1.5em}
\end{table}

\begin{table}[t]
    \tablestyle{8pt}{1.1}
    \begin{tabular}{@{}l|llll@{}}
        prop. strategy & \multicolumn{1}{c}{AP$^\text{box}$} & \# params & train mem & test time \\
        \shline
        none  &  52.9 & 1.00$\times$ {\scriptsize (331M)} & 1.00$\times$ {\scriptsize (14.6G)} & 1.00$\times$ {\scriptsize (88ms)} \\ 
        \hline  
        4 conv (bottleneck) & \res{54.6}{+1.7} & 1.04$\times$ & 1.05$\times$ & 1.04$\times$ \\
        4 global & \res{54.6}{+1.7} & 1.00$\times$ & 1.39$\times$ & 1.16$\times$ \\
        24 global & \res{55.1}{+2.2} & 1.00$\times$ & 3.34$\times$$^\dagger$ & 1.86$\times$ \\
    \end{tabular}
    \vspace{.5em}
    \caption{\textbf{Practical performance of backbone adaptation strategies}. The backbone is \mbox{ViT-L}. The training memory (per GPU) is benchmarked with a batch size of 1. The testing time (per image) is benchmarked on an A100 GPU. {$^\dagger$: This 3.34$\times$ memory (49G) is estimated as if the same training implementation could be used, which is not practical and requires special memory optimization that all together slows down training by 2.2$\times$ \vs the baseline.}
    \label{tab:complexity}
    }
\vspace{-1.5em}
\end{table}

In Table~\ref{tab:backbone_ablation:prop}, we compare our global and convolutional propagation strategies \vs the no propagation baseline. They have a gain of 1.7 and 1.9 over the baseline. We also compare with the ``shifted window" (Swin \cite{Liu2021}) strategy, in which the window grid is shifted by a half-window size for every other block. The shifted window variant has a 1.1 gain over the baseline, but is worse than ours.
Note that here we focus only on the ``shifted window" aspect of Swin \cite{Liu2021}: the backbone is still a plain ViT, adapted to shifted window attention only during fine-tuning; it is \textit{not} the Swin architecture, which we will compare to later.

Table~\ref{tab:backbone_ablation:conv_type} compares different types of residual blocks for convolutional propagation. We study the basic (two 3$\times$3) \cite{He2016}, bottleneck (1$\times$1$\rightarrow$3$\times$3$\rightarrow$1$\times$1) \cite{He2016}, and a na\"ive block that has one 3$\times$3 convolution. They all improve over the baseline, while the specific block design makes only marginal differences. Interestingly, even though convolution is a local operation if its receptive field covers two adjacent windows, it is sufficient in principle to connect all pixels of the two windows. This connectivity is thanks to the self-attention in both windows in the succeeding blocks. This may explain why it can perform as well as global propagation.

In Table~\ref{tab:backbone_ablation:block_place} we study where cross-window propagation should be located in the backbone.
By default 4 global propagation blocks are placed \textit{evenly}. We compare with placing them in the first or last 4 blocks instead. Interestingly, performing propagation in the last 4 blocks is nearly as good as even placement. 
This is in line with the observation in \cite{Dosovitskiy2021} that ViT has longer attention distance in later blocks and is more localized in earlier ones.
In contrast, performing propagation only in the first 4 blocks shows no gain: in this case, there is no propagation across windows in the backbone after these 4 blocks. This again demonstrates that propagation across windows is helpful.

\begin{table}[t]
    \tablestyle{8pt}{1.1}
    \begin{tabular}{@{}l|ll|ll@{}}
     & \multicolumn{2}{c|}{ViT-B} & \multicolumn{2}{c}{ViT-L} \vspace{-.5em} \\
    pre-train & \multicolumn{1}{c}{\scriptsize \boxAP} & \multicolumn{1}{c|}{\scriptsize \maskAP} & \multicolumn{1}{c}{\scriptsize \boxAP} & \multicolumn{1}{c}{\scriptsize \maskAP} \\
    \shline
    none (random init.) &  48.1 & 42.6 & 50.0 & 44.2 \\
    \hline
IN-1K, supervised & \bad{47.6}{0.5} & \bad{42.4}{0.2} & \bad{49.6}{0.4} & \bad{43.8}{0.4}  \\
IN-21K, supervised & \bad{47.8}{0.3} & \res{42.6}{+0.0} & \res{50.6}{+0.6} & \res{44.8}{+0.6} \\ 
IN-1K, MAE & \res{\textbf{51.2}}{+3.1} & \res{\textbf{45.5}}{+2.9} & \res{\textbf{54.6}}{+4.6} & \res{\textbf{48.6}}{+4.4} \\
    \end{tabular}

    \vspace{.5em}
    \caption{\textbf{Ablation on pre-training strategies} with plain ViT backbones using {Mask R-CNN} evaluated on COCO.
    \label{tab:pre-training}
    }
\vspace{-2em}
\end{table}

Table~\ref{tab:backbone_ablation:block_num} compares the number of global propagation blocks to use. Even using just 2 blocks achieves good accuracy and clearly outperforms the baseline. For comprehensiveness, we also report a variant where all 24 blocks in ViT-L use global attention. This has a marginal gain of 0.5 points over our 4-block default, while its training requires special memory optimization (we use memory checkpointing \cite{Chen2016}). This requirement makes scaling to larger models (like \mbox{ViT-H}) impractical. Our solution of window attention plus a few propagation blocks offers a practical, high-performing tradeoff. 

We benchmark this tradeoff in Table~\ref{tab:complexity}.
Using 4 propagation blocks gives a good trade-off. Convolutional propagation is the most practical, increasing memory and time by merely $\leq$5\%, at a small cost of 4\% more parameters.
 Global propagation with 4 blocks is also feasible and does not increase the model size. Global self-attention in all 24 blocks is not practical.

In sum, Table~\ref{tab:backbone_ablations} shows that various forms of propagation are helpful, while \textit{we can keep using window attention in most or all blocks}.
 Importantly, all these architecture adaptations are performed only during fine-tuning time; they do not require a redesign of the pre-training architecture.

\paragraph{Masked Autoencoders provide strong pre-trained backbones.} Table~\ref{tab:pre-training} compares backbone pre-training strategies. Supervised pre-training on IN-1K is slightly worse than no pre-training, similar to the observation in~\cite{Ghiasi2021}. Supervised pre-training on IN-21K is marginally better for ViT-L.

In contrast, MAE \cite{He2021} pre-training on IN-1K (without labels) shows massive gains, increasing \boxAP by 3.1 for ViT-B and 4.6 for ViT-L. 
We hypothesize that the vanilla ViT \cite{Dosovitskiy2021}, with fewer inductive biases, may require higher-capacity to learn translation and scale equivariant features, while higher-capacity models are prone to heavier overfitting. MAE pre-training can help to relieve this problem. We discuss more about MAE in context next.

\subsection{Comparisons with Hierarchical Backbones} \label{subsec:vs_hier}

Modern detection systems involve many implementation details and subtleties. To focus on comparing backbones under as fair conditions as possible, we incorporate the Swin \cite{Liu2021} and MViTv2 \cite{Li2021a} backbones into our implementation.

\paragraph{Settings.} We use the same implementation of Mask R-CNN \cite{He2017} and Cascade Mask R-CNN \cite{Cai2019} for all ViT, Swin, and MViTv2 backbones. We use FPN for the hierarchical backbones of Swin/MViTv2. We search for optimal hyper-parameters separately for each backbone (see the appendix). Our Swin results are better than their counterparts in the original paper;\footnotemark~our MViTv2 results are better than or on par with those reported in \cite{Li2021a}.

\footnotetext{For example, Swin-B (IN-1K, Cascade Mask R-CNN) has 51.9 \boxAP reported in the official repo. This result in our implementation is 52.7.}

Following the original papers \cite{Liu2021,Li2021a}, Swin and MViTv2 both use {relative position biases} \cite{Raffel2020}. For a fairer comparison, here we also adopt relative position biases in our ViT backbones as per \cite{Li2021a}, but \textit{only} during fine-tuning, not affecting pre-training. 
This addition improves AP by $\app$1 point. Note that our ablations in Sec.~\ref{subsec:ablation} are \textit{without} relative position biases.

\definecolor{deemph}{gray}{0.7}
\newcolumntype{g}{>{\color{deemph}}r}
\begin{table}[t]
    \centering
    \tablestyle{2pt}{1.05}
    \begin{tabular}{@{}y{48}x{48}|x{32}x{32}|x{32}x{32}}
		& & 
		\multicolumn{2}{c|}{\scriptsize Mask R-CNN} & 
		\multicolumn{2}{c}{\scriptsize Cascade Mask R-CNN}
		\\
        backbone &  pre-train &
        \multicolumn{1}{c}{\scriptsize \boxAP} & \multicolumn{1}{c|}{\scriptsize \maskAP} &
        \multicolumn{1}{c}{\scriptsize \boxAP} & \multicolumn{1}{c}{\scriptsize \maskAP} \\
        \shline
        \multicolumn{3}{@{}l}{\emph{hierarchical-backbone detectors:}} \\
        \hline
        Swin-B & 21K, sup & 51.4 & 45.4 &
        54.0 & 46.5 \\
        Swin-L & 21K, sup & 52.4 & 46.2 &
        54.8 & 47.3 \\
        \hline
        MViTv2-B & 21K, sup & 53.1 & 47.4 & 
        55.6 & 48.1 \\ 
        MViTv2-L & 21K, sup & 53.6 & 47.5 & 
        55.7 & 48.3 \\
        MViTv2-H & 21K, sup & 54.1 & 47.7 & 
        55.8 & 48.3 \\
        \hline
        \multicolumn{3}{@{}l}{\emph{our plain-backbone detectors:}} \\
        \hline
        ViT-B & 1K, {\scriptsize MAE} & 51.6 & 45.9 &
        54.0 & 46.7 \\
        ViT-L  & 1K, {\scriptsize MAE} & 55.6 & 49.2 &
        57.6 & 49.8 \\
        ViT-H & 1K, {\scriptsize MAE} & \textbf{56.7} & \textbf{50.1} &
        \textbf{58.7} & \textbf{50.9} \\
    \end{tabular}
    \vspace{.5em}
    \caption{\textbf{Comparisons of plain \vs hierarchical backbones} using Mask R-CNN \cite{He2017} and Cascade Mask R-CNN \cite{Cai2019} on COCO. Tradeoffs are plotted in Figure~\ref{fig:tradeoff}. All entries are implemented and run by us to align low-level details.
    \label{tab:coco_results}
    }
    \vspace{-.5em}
\end{table}
\begin{figure}[t]
\vspace{-1em}
\newcommand{\sz}{0.295}
\makebox[\textwidth][c]{
\begin{minipage}{1.25\linewidth}  %
\includegraphics[height=\sz\linewidth,trim={0 0 0 0},clip]{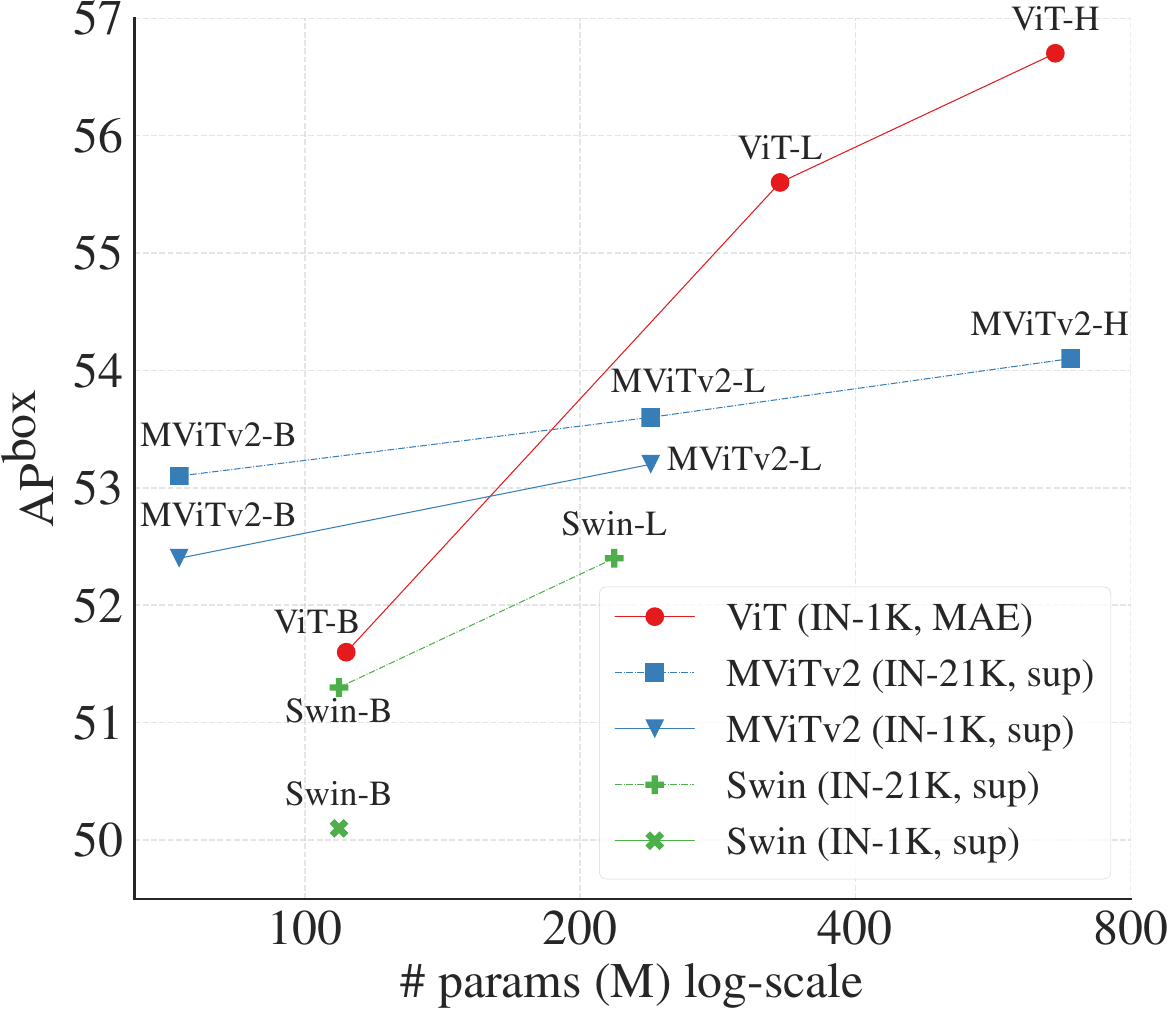}
\includegraphics[height=\sz\linewidth,trim={34px 0 0 0},clip]{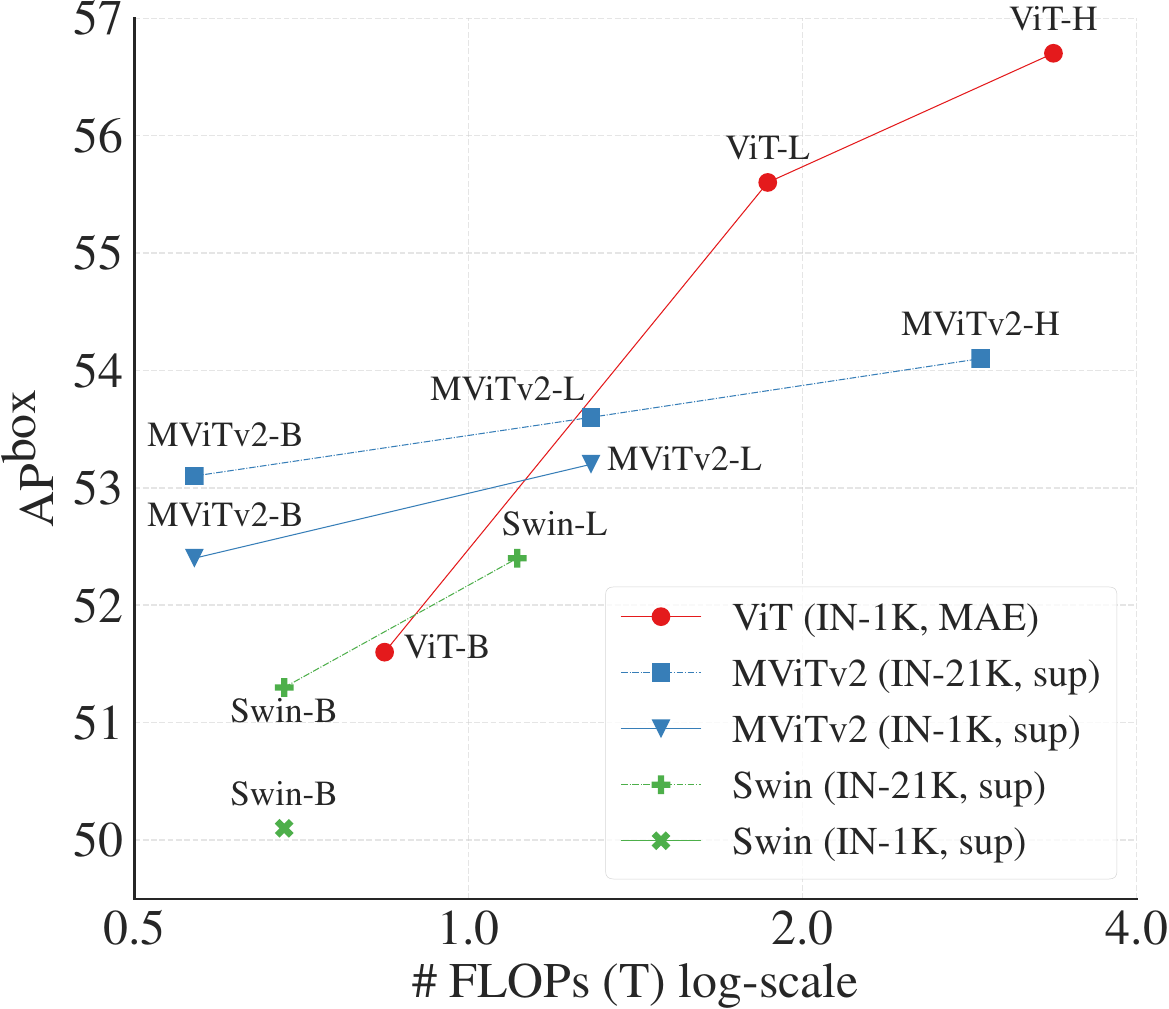}
\includegraphics[height=\sz\linewidth,trim={34px 0 0 0},clip]{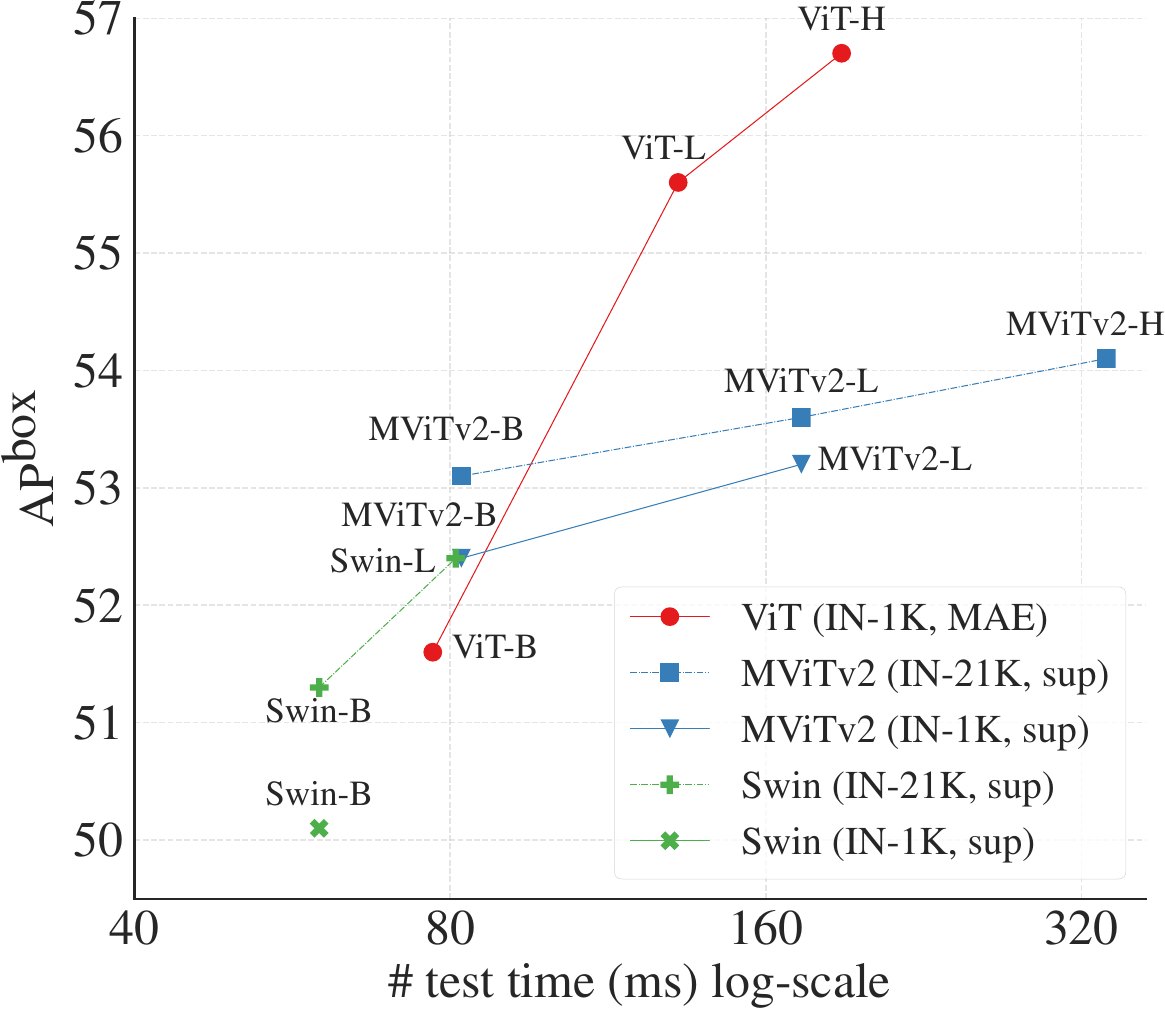}
\end{minipage}
}
\vspace{-.5em}
\caption{Tradeoffs of accuracy \vs model sizes (left), FLOPs (middle), and wall-clock testing time (right).
All entries are implemented and run by us to align low-level details.
Swin \cite{Liu2021} and MViTv2 \cite{Li2021a} are pre-trained on IN-1K/21K with supervision. The ViT models are pre-trained using MAE \cite{He2021} on IN-1K.
Here the detector head is \mbox{Mask R-CNN}; similar trends are observed for Cascade Mask R-CNN and one-stage detector RetinaNet (Figure~\ref{fig:retinanet_tradeoff} in the appendix). Detailed numbers are in the appendix (Table~\ref{app:tab:coco_full_results}).
\label{fig:tradeoff}
}
\vspace{-1.5em}
\end{figure}

\paragraph{Results and analysis.} Table~\ref{tab:coco_results} shows the comparisons.
Figure~\ref{fig:tradeoff} plots the tradeoffs.
The comparisons here involve two factors: the backbone and the pre-training strategy. Our plain-backbone detector, combined with MAE pre-training, presents \textit{better scaling behavior}. When the models are large, our method outperforms the hierarchical counterparts of Swin/MViTv2, including those using IN-21K supervised pre-training. Our result with ViT-H is 2.6 better than that with MViTv2-H. Moreover, the plain ViT has a \textit{better} wall-clock performance (Figure~\ref{fig:tradeoff} right, see ViT-H \vs MViTv2-H), as the {simpler} blocks are more hardware-friendly.

We are also curious about the influence of MAE on hierarchical backbones. This is largely beyond the scope of this paper, as it involves finding good training recipes for hierarchical backbones with MAE.
To provide some insight, we implement a na\"ive extension of MAE with the MViTv2 backbone (see the appendix). We observe that MViTv2-L with this MAE pre-training on IN-1K is 1.3 better than that with IN-21K supervised pre-training (54.9 \vs 53.6 \boxAP). As a comparison, this gap is 4 points for our plain-backbone detector (Table~\ref{tab:pre-training}). This shows that the plain ViT backbone may benefit \textit{more} from MAE pre-training than the hierarchical backbone, suggesting that the lack of inductive biases on scales could be compensated by the self-supervised training of MAE.
While it is an interesting future topic on improving hierarchical backbones with MAE pre-training, our plain-backbone detector enables us to use the \textit{readily available} ViT backbones from MAE to achieve strong results.

We also note that hierarchical backbones in general involve \textit{enhanced} self-attention block designs. Examples include the shifted window attention in Swin \cite{Liu2021} and pooling attention in MViT v1/v2 \cite{Fan2021,Li2021a}. These block designs, if applied to plain backbones, may also improve accuracy and parameter-efficiency.
While this may put our competitors at an advantage, our method is still competitive without these enhancements.

\newcommand{\deemph}[1]{{\color{deemph}#1}}
\newcolumntype{k}{>{\color{deemph}}c}
\begin{table}[t]
    \tablestyle{6pt}{1.02}
    \begin{tabular}{@{}lll|cc|kk@{}}
    & & & \multicolumn{2}{c|}{\scriptsize single-scale test} & \multicolumn{2}{k}{\scriptsize multi-scale test} \\
    method & framework &  pre-train & {\boxAP} & {\maskAP} & {\boxAP} & {\maskAP} \\
    \shline
    \multicolumn{3}{@{}l}{\emph{hierarchical-backbone detectors:}} \\
    \hline
    ~Swin-L~\cite{Liu2021} & HTC++ & 21K, sup & 57.1 & 49.5 & 58.0 & 50.4 \\
    ~MViTv2-L~\cite{Li2021a} & Cascade & 21K, sup & 56.9 & 48.6 & 58.7 & 50.5 \\
    ~MViTv2-H~\cite{Li2021a} & Cascade & 21K, sup & 57.1 & 48.8 & 58.4 & 50.1 \\
    ~CBNetV2~\cite{Liang2021}$^\dagger$ & HTC & 21K, sup & 59.1 & 51.0 & 59.6 & 51.8 \\
    ~SwinV2-L~\cite{Liu2021a} & HTC++ & 21K, sup & 58.9 & 51.2 & 60.2 & 52.1\\
    \hline
    \multicolumn{3}{@{}l}{\emph{plain-backbone detectors:}} \\
    \hline
    ~UViT-S~\cite{Chen2021b} & Cascade & 1K, sup & 51.9 & 44.5 & - & - \\
    ~UViT-B~\cite{Chen2021b} & Cascade & 1K, sup & 52.5 & 44.8 & - & - \\
    ~\textbf{ViTDet}, ViT-B & Cascade &  1K, {\scriptsize MAE} & 56.0 & 48.0 & 57.3 & 49.4 \\
    ~\textbf{ViTDet}, ViT-L & Cascade &  1K, {\scriptsize MAE} & 59.6 & 51.1 & 60.4 & 52.2 \\
    ~\textbf{ViTDet}, ViT-H & Cascade &  1K, {\scriptsize MAE} & \textbf{60.4} & \textbf{52.0} & \textbf{61.3} & \textbf{53.1} \\
    \end{tabular}
    \vspace{.5em}
    \caption{\textbf{System-level comparisons with the leading results on COCO} reported by the original papers. 
    The detection framework is Cascade Mask R-CNN~\cite{Cai2019} (denoted as ``Cascade"), Hybrid~Task~Cascade (HTC)~\cite{Chen2019}, or its extension (HTC++~\cite{Liu2021}).
    Here we compare results that use ImageNet data (1K or 21K); better results are reported in \cite{Liu2021a,Dai2021} using extra data.
    $^\dagger$: \cite{Liang2021} combines two Swin-L backbones.
    \label{tab:coco_system_results} 
    }
\vspace{-2em}
\end{table}

\subsection{Comparisons with Previous Systems} \label{subsec:vs_prev}

Next we provide \textit{system-level} comparisons with the leading results reported in previous papers. We refer to our system as \textbf{ViTDet}, \ie, ViT Detector, aiming at the usage of a ViT backbone for detection. Since these comparisons are system-level, the methods use a variety of different techniques. While we make efforts to balance the comparisons (as noted below), making a perfectly controlled comparison is infeasible in general; our goal, instead, is to situate our method in the context of current leading methods.

\paragraph{Comparisons on COCO.} Table~\ref{tab:coco_system_results} reports the system-level comparisons on COCO.
For a fairer comparison, here we make two changes following our competitors: we adopt soft-nms~\cite{Bodla2017} as is used by all competitors \cite{Liu2021,Li2021a,Liang2021,Liu2021a} in this table and increase the input size (from 1024 to 1280) following \cite{Liang2021,Liu2021a}. We note that we do \textit{not} use these improvements in previous ablations. As in the previous subsection (Sec.~\ref{subsec:vs_prev}), we use relative position biases here.

The leading systems thus far are all based on hierarchical backbones (Table~\ref{tab:coco_system_results}). For the first time, we show that a \textit{plain-backbone} detector can achieve highly accurate results on COCO and can compete with the leading systems.

We also compare with UViT~\cite{Chen2021b} which is a recent plain-backbone detection method. As discussed in Sec.~\ref{sec:related}, UViT and our work have different focuses. UViT aims at designing a new plain backbone that is good for detection, while our goal here is to support general-purpose ViT backbones including the original ones in \cite{Dosovitskiy2021}.
Despite the different focuses, both UViT and our work suggest that plain-backbone detection is a promising direction with strong potential.

\paragraph{Comparisons on LVIS.} We further report system-level comparisons on the LVIS dataset \cite{Gupta2019}. LVIS contains \app2M high-quality instance segmentation annotations for 1203 classes that exhibit a natural, long-tailed object distribution. Unlike COCO, the class distribution is heavily imbalanced and many classes have very few (\eg, $<$10) training examples.

We follow the same model and training details as used for the COCO system-level comparison plus two common LVIS practices: we use the federated loss from \cite{Zhou2021} and sample images with repeat factor sampling~\cite{Gupta2019}.
We fine-tune for 100 epochs on the v1 \textsf{train} split.

Table~\ref{tab:lvis_results} shows the results on the v1 \textsf{val} split. Our plain-backbone detector achieves competitive performance \vs previous leading results that all use hierarchical backbones. Ours is 5.0 points higher than the 2021 competition winner's ``strong baseline" \cite{Fu2021} (48.1 \vs 43.1 \maskAP), which uses HTC with CBNetV2~\cite{Liang2021} that combines two Swin-L backbones. A special issue in LVIS is on the long-tailed distribution, which is beyond the scope of our study. Techniques dedicated to this issue, \eg, using CLIP \cite{Radford2021} text embeddings or other advancements from \cite{Fu2021}, can largely increase AP on the rare classes (AP$^\text{mask}_\text{rare}$) and thus improve overall AP. These are orthogonal to our method and could be complementary. Nevertheless, our results on LVIS again suggest that plain-backbone detectors can compete with hierarchical ones.
 
\begin{table}[t]
    \tablestyle{6pt}{1.02}
    \begin{tabular}{@{}ll|ccc@{}}
    method  & pre-train & AP$^\text{mask}$ & AP$^\text{mask}_\text{rare}$ & AP$^\text{box}$\\
    \shline
    \multicolumn{3}{@{}l}{\emph{hierarchical-backbone detectors:}} \\
    \hline
    ~Copy-Paste~\cite{Ghiasi2021}, Eff-B7 FPN & none (random init) & 36.0 & 29.7 & 39.2 \\
    ~Detic~\cite{Zhou2022}, Swin-B & 21K, sup; CLIP & 41.7 &  41.7  & - \\
    ~competition winner 2021~\cite{Fu2021} baseline, $^\dagger$ & 21K, sup & 43.1 & 34.3 & - \\
    ~competition winner 2021~\cite{Fu2021} full, $^\dagger$ & {21K, sup} & \textbf{49.2} & \textbf{45.4}  & - \\
    \hline
    \multicolumn{3}{@{}l}{\emph{plain-backbone detectors:}} \\
    \hline
    ~\textbf{ViTDet}, ViT-L & 1K, {\scriptsize MAE} & 46.0 & 34.3  &  51.2\\
    ~\textbf{ViTDet}, ViT-H & 1K, {\scriptsize MAE} & 48.1 & 36.9 &  53.4 \\
    \end{tabular}
    \vspace{.5em}
    \caption{\textbf{System-level comparisons with the leading results on LVIS} (v1 \textsf{val}) reported by the original papers. All results are without test-time augmentation.
    Detic \cite{Zhou2022} uses pre-trained CLIP \cite{Radford2021} text embeddings.
    $^\dagger$: these entries use CBNetV2 \cite{Liang2021} that combines two \mbox{Swin-L} backbones.
     \label{tab:lvis_results}
    }
    \vspace{-2em}
\end{table}

\section{Conclusion}

Our exploration has demonstrated that \textit{plain-backbone detection is a promising research direction}. This methodology largely maintains the independence of the general-purpose backbones and the downstream task-specific designs---which had been the case for ConvNet-based research but not for Transformer-based research. We hope decoupling pre-training from fine-tuning is a methodology that will generally benefit the community. For example, in natural language processing (NLP), general-purpose pre-training (GPT \cite{Radford2018}, BERT \cite{Devlin2019}) has greatly advanced the field and has been supporting various downstream tasks. In this study, our plain-backbone detector has benefited from the readily available pre-trained models from MAE \cite{He2021}. We hope this methodology will also help bring the fields of computer vision and NLP closer.

\clearpage

\appendix

\section{Appendix}

\newcommand{\lr}{\emph{lr}\xspace}
\newcommand{\wtd}{\emph{wd}\xspace}
\newcommand{\drp}{\emph{dp}\xspace}
\newcommand{\expnum}[2]{{#1}\mathrm{e}^{#2}}

\renewcommand{\citeapp}{\cite}

\subsection{Additional Ablation Results}\label{app:sec:results}

~~
Table~\ref{app:tab:backbone_ablations_vit_b} is the ViT-B counterpart of Table~\ref{tab:backbone_ablations} on backbone adaptation. The observations are similar to that of ViT-L: comparing with the baseline using no propagation (``none"), various propagation strategies show good gains.

Table~\ref{app:tab:coco_full_results} presents Table~\ref{tab:coco_results} with additional details about FLOPs, parameters, and inference time, plotted in Figure~\ref{fig:tradeoff}.

Table~\ref{app:tab:pre-training-lvis} is the ablation on pre-training strategies for LVIS. Similar to Table~\ref{tab:pre-training}, MAE pre-training has large gains over supervised pre-training.

\begin{figure}[t]
    \newcommand{\sz}{0.295}
    \makebox[\textwidth][c]{
    \begin{minipage}{1.25\linewidth}  %
    \includegraphics[height=\sz\linewidth,trim={0 0 0 0},clip]{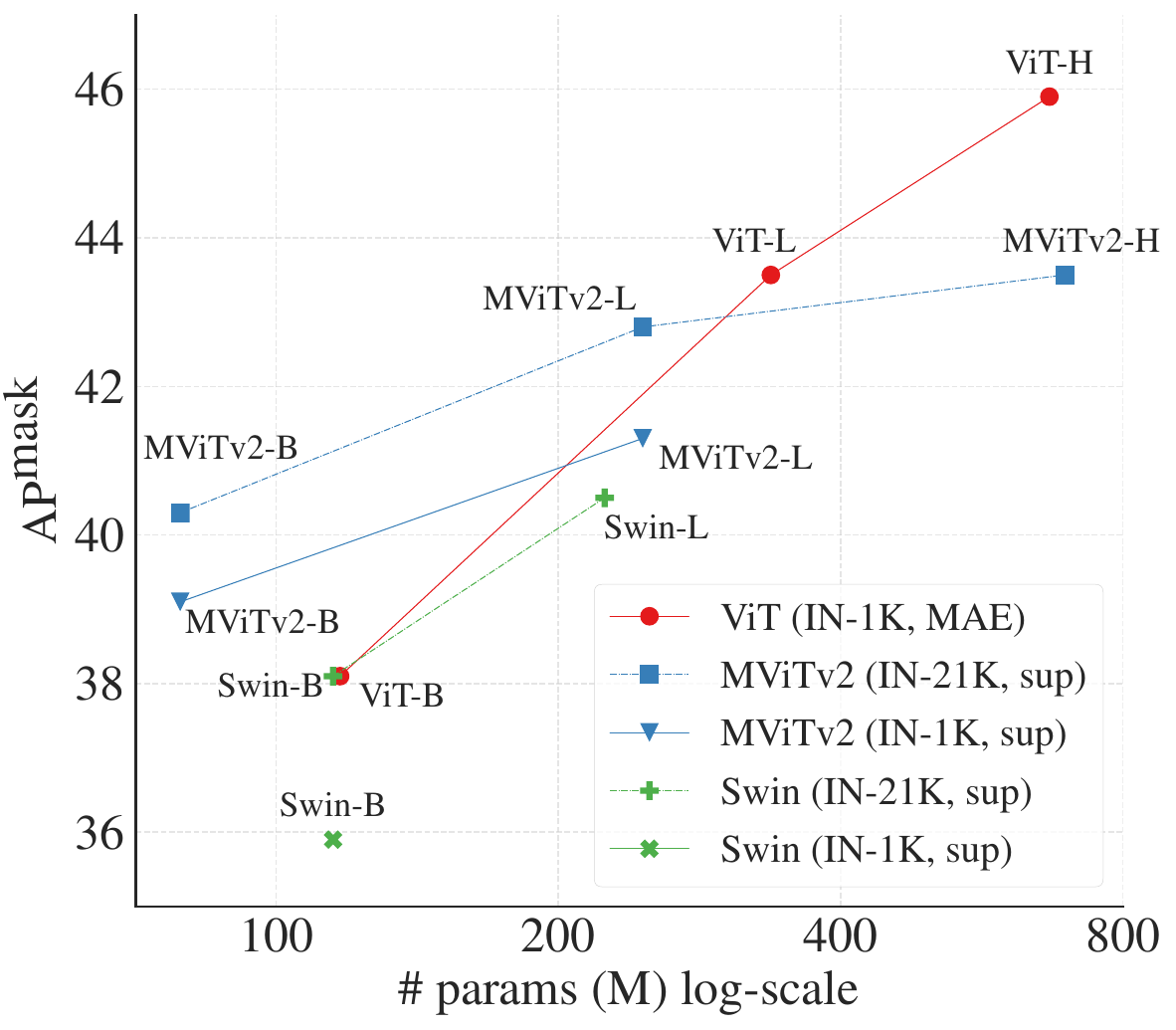}
    \includegraphics[height=\sz\linewidth,trim={34px 0 0 0},clip]{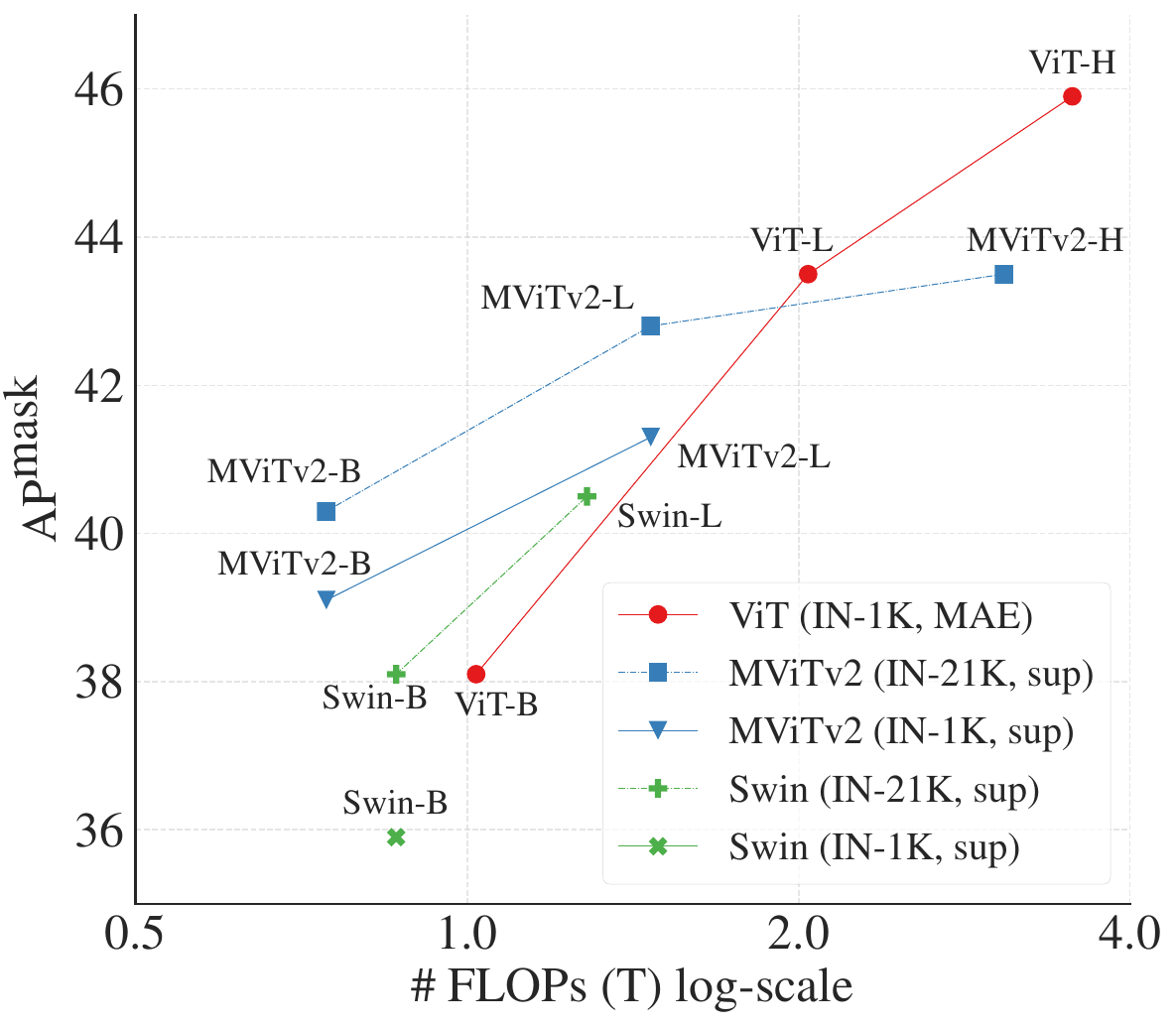}
    \includegraphics[height=\sz\linewidth,trim={34px 0 0 0},clip]{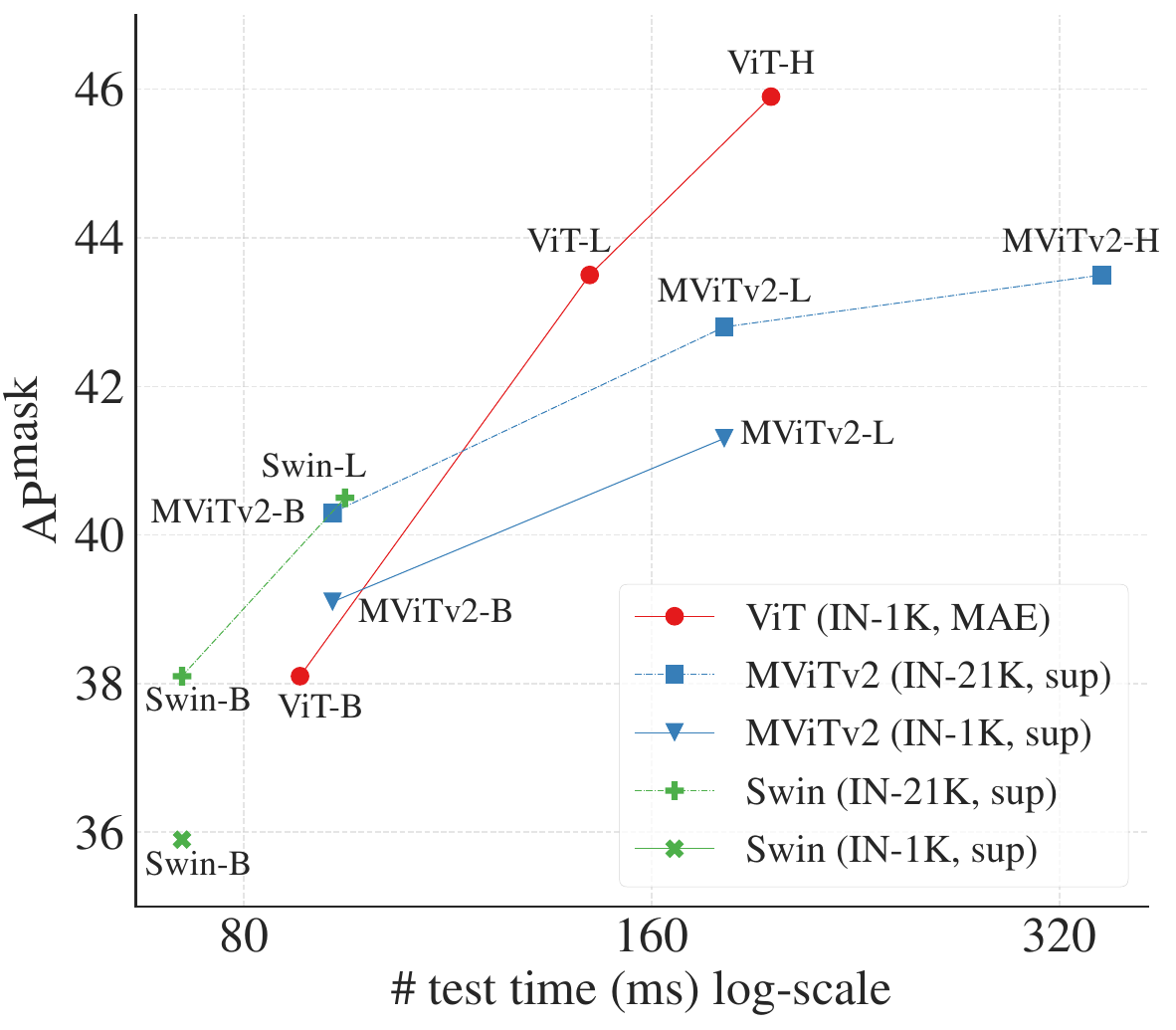}
    \end{minipage}
    }
    \vspace{-.5em}
    \caption{The LVIS counterpart of Figure~\ref{fig:tradeoff}.
    All entries are implemented and run by us to align low-level details.
    Here the detector head is \mbox{Mask R-CNN} (input resolution 1024; no soft-nms). The trends are similar to those in Figure~\ref{fig:tradeoff}, while IN-21K supervised pre-training has larger gains.
    \label{app:fig:tradeoff}
    }
\end{figure}

Figure~\ref{app:fig:tradeoff} is the LVIS counterpart of Figure~\ref{fig:tradeoff}. The trends are similar to those in COCO, while the gain of IN-21K supervised pre-training is larger because it significantly improves rare category detection in LVIS.

\begin{figure}[t]\centering
\includegraphics[width=0.6\linewidth,trim={0 0 0 0},clip]{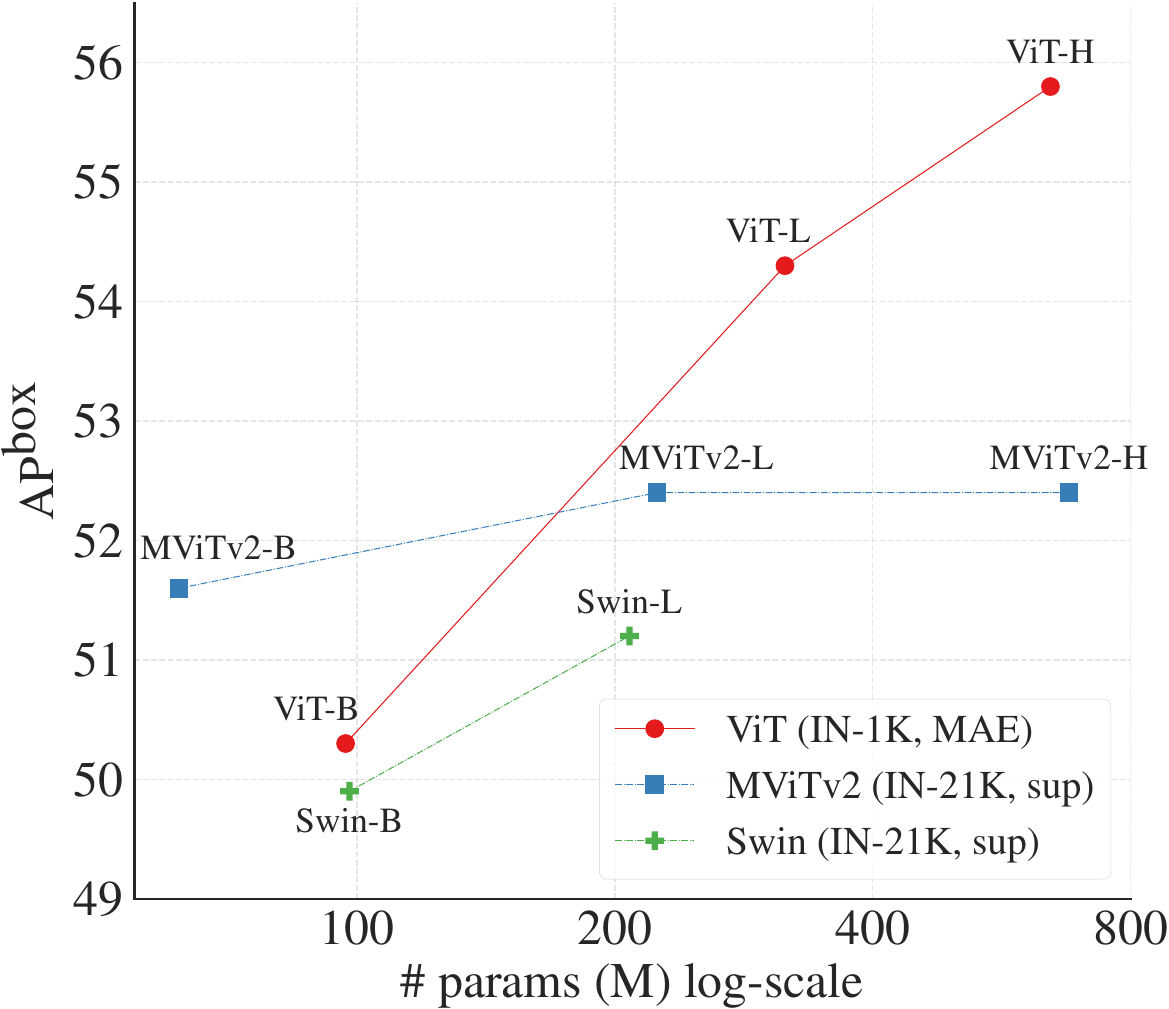}
\caption{The RetinaNet~\cite{Lin2017a} counterpart of Figure~\ref{fig:tradeoff}, showing the trade-off between accuracy and model size. We use the same Mask R-CNN training recipe (input resolution 1024; no soft-nms) and hyper-parameters for RetinaNet. The trends are similar to those in Figure~\ref{fig:tradeoff}.
\label{fig:retinanet_tradeoff}
}
\vspace{-1em}
\end{figure}

Figure~\ref{fig:retinanet_tradeoff} is the RetinaNet~\cite{Lin2017a} counterpart of Figure~\ref{fig:tradeoff}, showing the trade-off between accuracy and model size. Here, we evaluate ViTDet with a one-stage RetinaNet~\cite{Lin2017a} detector head and compare it to using Swin and MViTv2 as hierarchical backbones, all without hyper-parameter tuning. Compared to using Mask R-CNN and Cascade R-CNN (Table~\ref{tab:coco_results} and Figure~\ref{fig:tradeoff}), we observe similar trends with RetinaNet. In particular, our plain-backbone detector presents \emph{better scaling behavior} (\eg. ViT-H gains \textbf{+3.4} \boxAP over MViTv2-H). These results suggest that the proposed training recipe transfers well to different detectors and that our proposed plain backbone adaptations are general and can likely work with even more detection architectures.

\subsection{Implementation Details} \label{app:sec:details}

\paragraph{Architectures.}
We build a simple feature pyramid of scales $\{\frac{1}{32}, \frac{1}{16}, \frac{1}{8}, \frac{1}{4}\}$ (see Sec.~\ref{sec:method}). The $\frac{1}{32}$ scale is built by stride-2 2$\times$2 max pooling (average pooling or convolution works similarly). The $\frac{1}{16}$ scale simply uses the ViT's final feature map. Scale $\frac{1}{8}$ (or $\frac{1}{4}$) is built by one (or two) 2$\times$2 deconvolution layer(s) with stride=2. In the $\frac{1}{4}$ scale case, the first deconvolution is followed by LayerNorm (LN) \citeapp{Ba2016} and GeLU \citeapp{Hendrycks2016}. Then for each pyramid level, we apply a 1$\times$1 convolution with LN to reduce dimension to 256 and then a 3$\times$3 convolution also with LN, similar to the per-level processing of FPN \cite{Lin2017}.

We study three detection frameworks: Mask R-CNN~\cite{He2017}, Cascade Mask R-CNN~\cite{Cai2019} and RetinaNet~\cite{Lin2017a}. For Mask R-CNN and Cascade Mask R-CNN, we incorporate some common best practices developed since they~\cite{He2017,Cai2019} were presented years ago. We use 2 hidden convolution layers for the RPN and 4 hidden convolution layers for the RoI heads as per \citeapp{Wu2018}. These hidden convolution layers are followed by LN. For all three detection frameworks, We use the same detection implementation for both plain and hierarchical backbones.

We use a patch size of 16 for all ViT backbones. As ViT-H in \cite{Dosovitskiy2021} by default has a patch size of 14, after pre-training we interpolate the patch embedding filters from 14$\times$14$\times$3 to 16$\times$16$\times$3. 

\paragraph{Hyper-parameters for COCO.}
Our default training recipe is as follows (unless noted in context for ablation).
The input size is 1024$\times$1024, augmented during training by {large-scale jitter}~\cite{Ghiasi2021} with a scale range of $[0.1, 2.0]$. We use AdamW~\cite{Loshchilov2019} ($\beta_1, \beta_2{=}0.9, 0.999$) with step-wise learning rate decay. We use linear learning rate warm-up \cite{Goyal2017} for 250 iterations. The batch size is 64, distributed across 64 GPUs (1 image per GPU).

We search for the learning rate (\lr), weight decay (\wtd), drop path rate (\drp), and epochs, for each model size (B, L, H) and for each model type (ViT, Swin, MViTv2).
The hyper-parameters used are in Table~\ref{app:tab:hyper}. We also use a layer-wise \lr decay \citeapp{Clark2020}\cite{Bao2021} of 0.7/0.8/0.9 for ViT-B/L/H with MAE pre-training, which has a small gain of up to 0.3 AP; we have not seen this gain for hierarchical backbones or ViT with supervised pre-training.

\begin{table}[t]
\vspace{-1em}
\centering
\subfloat[
Window attention with various cross-window propagation strategies.
\label{app:tab:backbone_ablation:prop_vitb}
]{
\centering
\begin{minipage}{0.46\linewidth}{\begin{center}
\tablestyle{4pt}{1.1}
\begin{tabular}{@{}y{60}|y{42}y{42}l@{}}
   prop. strategy & \multicolumn{1}{c}{AP$^\text{box}$} & \multicolumn{1}{c}{AP$^\text{mask}$} \\
    \shline
    none & {48.9} & 43.9 \\ 
    \hline
    4 global blocks & \res{\textbf{51.2}}{+2.3} & \res{\textbf{45.5}}{+1.6} \\
    4 conv blocks & \res{{51.0}}{+2.1} & \res{{45.3}}{+1.4} \\
    shifted win. & \res{50.1}{+1.2} & \res{44.8}{+0.9} \\
\end{tabular}
\end{center}}\end{minipage}
}
\hspace{1em}
\subfloat[
Convolutional propagation with different residual block types (4 blocks).
\label{app:tab:backbone_ablation:conv_type_vitb}
]{
\begin{minipage}{0.46\linewidth}{\begin{center}
\tablestyle{4pt}{1.1}
\begin{tabular}{@{}y{40}|y{42}y{42}@{}}
    prop. conv &  \multicolumn{1}{c}{AP$^\text{box}$} & \multicolumn{1}{c}{AP$^\text{mask}$} \\
    \shline
    none & {48.9} & 43.9\\
    \hline
    na\"ive & \res{50.6}{+1.7} & \res{45.2}{+1.3} \\
    basic & \res{{50.7}}{+1.8} & \res{{45.2}}{+1.3}\\
    bottleneck & \res{\textbf{51.0}}{+2.1} & \res{\textbf{45.3}}{+1.4}\\ 
    \end{tabular}
\end{center}}\end{minipage}
}
\\
\subfloat[Locations of cross-window global propagation blocks.
\label{app:tab:backbone_ablation:block_place_vitb}
]{
\begin{minipage}{0.46\linewidth}{\begin{center}
\tablestyle{4pt}{1.1}
\begin{tabular}{@{}y{60}|y{42}y{42}@{}}
    prop. locations & \multicolumn{1}{c}{AP$^\text{box}$} & \multicolumn{1}{c}{AP$^\text{mask}$} \\
    \shline
    none & {48.9} & 43.9\\
    \hline
    first 4 blocks & \res{49.1}{+0.2} & \res{44.1}{+0.2}  \\
    last 4 blocks & \res{50.9}{+2.0} & \res{45.4}{+1.5} \\
    evenly 4 blocks & \res{\textbf{51.2}}{+2.3} & \res{\textbf{45.5}}{+1.6} \\
    \end{tabular}
\end{center}}\end{minipage}
}
\hspace{1em}
\subfloat[Number of global propagation blocks.
\label{app:tab:backbone_ablation:block_num_vitb}
]{
\begin{minipage}{0.46\linewidth}{\begin{center}
\tablestyle{4pt}{1.1}
\begin{tabular}{@{}y{40}|y{42}y{42}@{}}
    prop. blks &  \multicolumn{1}{c}{AP$^\text{box}$} & \multicolumn{1}{c}{AP$^\text{mask}$} \\
    \shline
    none & {48.9} & 43.9\\
    \hline
    2 & \res{50.7}{+1.8} & \res{45.2}{+1.3} \\
    4 & \res{\textbf{51.2}}{+2.3} & \res{\textbf{45.5}}{+1.6} \\
    12 & \res{{50.4}}{+1.5} & \res{{45.1}}{+1.2} \\ 
    \end{tabular}
\end{center}}\end{minipage}
}
\vspace{-.5em}
\caption{The ViT-B counterpart of Table~\ref{tab:backbone_ablations} (backbone adaptation).
}
\label{app:tab:backbone_ablations_vit_b}
\vspace{-1em}
\end{table}

\definecolor{deemph}{gray}{0.7}
\vspace{-.5em}
\begin{table*}[t]
    \centering
    \tablestyle{3pt}{1.05}
    \begin{tabular}{@{}l|l|ccrrr|ccrrr@{}}
		& & 
		\multicolumn{5}{c|}{\scriptsize Mask R-CNN} & 
		\multicolumn{5}{c}{\scriptsize Cascade Mask R-CNN}
		\\
        backbone &  pre-train &
        \multicolumn{1}{c}{\scriptsize \boxAP} & \multicolumn{1}{c}{\scriptsize \maskAP} &
        \scriptsize FLOPs & \scriptsize params & \scriptsize time & 
        \multicolumn{1}{c}{\scriptsize \boxAP} & \multicolumn{1}{c}{\scriptsize \maskAP} &
        \scriptsize FLOPs & \scriptsize params & \scriptsize time  \\
        \shline
        \multicolumn{5}{@{}l}{\emph{hierarchical-backbone detectors:}} \\
        \hline
        ~Swin-B & 1K, sup & 50.1 & 44.5 & 0.7T & 109M  & 60\scriptsize{ms} & %
        52.7 & 45.5 & 0.9T & 139M & 76\scriptsize{ms} \\ %
        ~Swin-B & 21K, sup & 51.4 & 45.4 & 0.7T & 109M & 60\scriptsize{ms} & %
        54.0 & 46.5 & 0.9T & 139M & 76\scriptsize{ms} \\ %
        ~Swin-L & 21K, sup & 52.4 & 46.2 & 1.1T & 218M & 81\scriptsize{ms} & %
        54.8 & 47.3 & 1.4T & 248M & 96\scriptsize{ms}  \\ %
        \hline
        ~MViTv2-B & 1K, sup & 52.4 & 46.7 & 0.6T & 73M & 82\scriptsize{ms} & %
        54.7 & 47.5 & 0.8T & 103M & 97\scriptsize{ms} \\  %
        ~MViTv2-L & 1K, sup & 53.2 & 47.1 & 1.3T & 239M & 173\scriptsize{ms} & %
        55.2 & 47.7 & 1.6T & 270M & 189\scriptsize{ms} \\ %
        ~MViTv2-B & 21K, sup & 53.1 & 47.4 & 0.6T & 73M & 82\scriptsize{ms} & %
        55.6 & 48.1 & 0.8T & 103M & 97\scriptsize{ms} \\  %
        ~MViTv2-L & 21K, sup & 53.6 & 47.5 & 1.3T & 239M & 173\scriptsize{ms} & %
        55.7 & 48.3 & 1.6T & 270M & 189\scriptsize{ms} \\ %
        ~MViTv2-H & 21K, sup & 54.1 & 47.7 & 2.9T & 688M & 338\scriptsize{ms} & %
        55.8 & 48.3 & 3.2T & 718M & 353\scriptsize{ms} \\ %
        \hline
        \multicolumn{5}{@{}l}{\emph{our plain-backbone detectors:}} \\
        \hline
        ~ViT-B & 1K, {\scriptsize MAE} & 51.6 & 45.9 & 0.8T & 111M & 77\scriptsize{ms} & %
        54.0 & 46.7 & 1.1T & 141M & 92\scriptsize{ms} \\
        ~ViT-L  & 1K, {\scriptsize MAE} & 55.6 & 49.2 & 1.9T & 331M & 132\scriptsize{ms} & %
        57.6 & 49.8 & 2.1T & 361M & 149\scriptsize{ms} \\ %
        ~ViT-H & 1K, {\scriptsize MAE} & \textbf{56.7} & \textbf{50.1} & 3.4T & 662M & 189\scriptsize{ms} & %
        \textbf{58.7} & \textbf{50.9} & 3.6T & 692M & 203\scriptsize{ms} \\ %
    \end{tabular}
    \vspace{.5em}
    \caption{Detailed measurements of Table~\ref{tab:coco_results} and Figure~\ref{fig:tradeoff}.
    \label{app:tab:coco_full_results}
    }
    \vspace{-1em}
    \end{table*}

\begin{table}[t]
    \tablestyle{8pt}{1.1}
    \begin{tabular}{@{}l|lll|lll@{}}
     & \multicolumn{3}{c|}{ViT-B} & \multicolumn{3}{c}{ViT-L} \vspace{-.5em} \\
    pre-train & \multicolumn{1}{c}{\scriptsize \boxAP} & \multicolumn{1}{c}{\scriptsize \maskAP} & \multicolumn{1}{c|}{\scriptsize \maskAPrare} & \multicolumn{1}{c}{\scriptsize \boxAP} & \multicolumn{1}{c}{\scriptsize \maskAP} & \multicolumn{1}{c}{\scriptsize \maskAPrare} \\
    \shline
IN-1K, supervised & 37.2 & 34.9 & 26.4 & 38.3 & 36.0 & 26.7 \\
IN-21K, supervised & 38.7 & 36.3 & 28.8 & 42.1 & 39.5 & 34.3 \\ 
IN-1K, MAE & \textbf{40.1} & \textbf{38.1} & \textbf{29.1} & \textbf{46.1} & \textbf{43.5} & \textbf{35.3} \\
    \end{tabular}
    \vspace{.5em}
    \caption{The LVIS counterpart of Table~\ref{tab:pre-training} (COCO pre-training ablation). The observations are similar to Table~\ref{tab:pre-training}: MAE pre-training has large gains over supervised pre-training. Here we also report rare category results. We observe that both IN-21K supervised and IN-1K MAE pre-training significantly improve \maskAPrare, especially for ViT-L. ({Mask R-CNN}, 1024 resolution, no soft-nms.)
    \label{app:tab:pre-training-lvis}
    }
\vspace{-1em}
\end{table}

\begin{table}[t]
    \tablestyle{6pt}{1.05}
    \begin{tabular}{llcccc}
    	backbone & pre-train & \lr & \wtd & \drp & epochs \\
    	\shline
    	ViT-B/L & none & $\expnum{1.6}{-4}$ & 0.2 & 0.1/0.4 & 300/200 \\
    	ViT-B/L & supervised & $\expnum{8}{-5}$ & 0.1 & 0.1/0.4 & 50 \\
    	ViT-B/L/H & MAE & $\expnum{1}{-4}$ & 0.1 & 0.1/0.4/0.5 & 100/100/75 \\
    	\hline
    	Swin-B/L & supervised & $\expnum{1}{-4}$/$\expnum{8}{-5}$ & 0.05 & 0.3 & 50 \\
    	MViTv2-B/L/H & supervised & $\expnum{8}{-5}$ & 0.1 & 0.4/0.5/0.6 & 100/50/36
    \end{tabular}
    \vspace{.5em}
    \caption{Hyper-parameters for COCO. Multiple values in a cell are for different model sizes. The epochs are chosen such that training longer starts to overfit.
     \label{app:tab:hyper}
    }
\end{table}

\paragraph{Hyper-parameters for LVIS.}
Our LVIS experiments in Table~\ref{tab:lvis_results} follow the COCO settings in Table~\ref{tab:coco_results}.
For LVIS, we set \lr = $\expnum{2}{-4}$/$\expnum{1}{-4}$ (ViT-L/H), \wtd = 0.1, and \drp = 0.4. We fine-tune for 100 epochs. We use a test score threshold of 0.02 (smaller values did not help) and repeat factor sampling ($t = 0.001$)~\cite{Gupta2019}. We output $\le$ 300 detections per image following \cite{Gupta2019} (\vs COCO's default 100).

\paragraph{MAE for hierarchical backbones.}

We implement a na\"ive extension of MAE pre-training~\cite{He2021} for the hierarchical backbone ablation (Sec.~\ref{subsec:vs_hier}).
MAE enjoys the efficiency benefit from plain ViT by skipping the encoder mask token \cite{He2021}. Extending this strategy to hierarchical backbones is beyond the scope of this paper. Instead, we adopt a straightforward solution in which we do not skip the encoder mask token (similar to \cite{Devlin2019}), at the cost of slower training.
We use normalized pixels as the MAE reconstruction target \cite{He2021} and set the decoder depth as 2. 

\vspace{1em}
\paragraph{Acknowledgement.} We would like to acknowledge Xinlei Chen, Saining Xie, Piotr Doll\'ar, and Christoph Feichtenhofer for discussions and support.

\bibliographystyle{ieee_fullname}
\bibliography{plain_det}

\end{document}